# A New Clustering Algorithm Based Upon Flocking On Complex Network

Qiang Li, Yan He, Jing-ping Jiang

College of Electrical Engineering, Zhejiang University,
Hang Zhou, Zhejiang, 310027, China

August 23, 2018

**Abstract**

We have proposed a model based upon flocking on a complex network, and then developed two clustering algorithms on the basis of it. In the algorithms, firstly a $k$-nearest neighbor (knn) graph as a weighted and directed graph is produced among all data points in a dataset each of which is regarded as an agent who can move in space, and then a time-varying complex network is created by adding long-range links for each data point. Furthermore, each data point is not only acted by its $k$ nearest neighbors but also $r$ long-range neighbors through fields established in space by them together, so it will take a step along the direction of the vector sum of all fields. It is more important that these long-range links provides some hidden information for each data point when it moves and at the same time accelerate its speed converging to a center. As they move in space according to the proposed model, data points that belong to the same class are located at a same position gradually, whereas those that belong to different classes are away from one another. Consequently, the experimental results have demonstrated that data points in datasets are clustered reasonably and efficiently, and the rates of convergence of clustering algorithms are fast enough. Moreover, the comparison with other algorithms also provides an indication of the effectiveness of the proposed approach.

## 1 Introduction

Data clustering is a widely investigated problem in Pattern Recognition. For the past forty years, a lot of excellent algorithms for clustering have been presented from those that put the emphasis on cluster centers and boundaries, say, $K$-means [1], support vector clustering (SVC) [2], to current particle swarm optimization (PSO) based [3], ant-based [4], and flocking-based [5] algorithms for clustering. Observing the history of clustering algorithms, we can notice that a significant change has been made, which may be considered as two stages. First, with fixed data points, we utilized various functions to find complex curve planes in order to cluster or classify data points; second, till the past few years, some pioneers thought about that why not those data points could move in themselves, just like agents or whatever, and collect together automatically. Therefore, following their ideas, they create a few exciting algorithms [3, 4, 5], in which data points moves in a whole space according to certain simple local rules preset in advance.

Flocking is a form of collective behavior among animals like birds, bees and fishes, which is to realize a group objective by interacting between individuals [6]. In the last ten years, many researchers with different backgrounds, ranging from physics, biology to computer sciences and sociology, are involved in this field in order to explore the mechanism of emergence of flocking with local interactions [7]. Certainly, flocking is also widely used in engineering applications, for example, self-assembly of connected mobile networks, massive distributed sensing using mobile sensor networks in an environment, etc. In particularly, flocking is applied to perform military missions as well, such as reconnaissance, surveillance, and combat using cooperative unmanned aerial vehicles.

To the best of our knowledge, flocking has begun to become an emerging method applied to the problem of data clustering. So, if data points for clustering are considered as a flock of agents who can move in space by local interactions, then could they collect together as separating parts automatically like the emergence of flocking? This is the question that we attempt to answer in this paper. In the proposed algorithms, the relationship among data points is represented by a time-varying complex network, on which data points interacts with its neighbors by a local potential function. Furthermore, each data point takes one step proportional to the magnitude of actions that it experiences along the direction of the actions. As data points move in space constantly, we can observe that they may gather together gradually and form some clusters automatically at last in the experiments. The remainder of this paper is organized as follows: Section 2 introduces some concepts and important parameters about the complex network theory briefly, and then reviews some related work about clustering algorithms based on flocking. Section 3 elaborates the proposed model of flocking on a complex network. Section 4 describes two clustering algorithms based on the model, in which the effects of long-range connections are analyzed in detail. Section 5 discusses the relation between the number of clusters and the number of nearest neighbors, and the rates of convergence of two algorithms. Section 6 introduces those datasets used in the experiments briefly, and then compares experimental results of the proposed algorithms with other clustering algorithms. The conclusion is given in Section 7.

# 2 Related work

## 2.1 Complex Network

There exist various networks or structures in the real world that we live [8], for example, the topology of food webs [9], electrical power grids, cellular and metabolic networks [10], the World-Wide Web [11], the neural network of the nematode worm [12], coauthorship and citation networks of scientists [13][14]. How to describe these networks in the real world is an issue that has puzzled researchers for about two hundred years. In the first one hundred years or so, the regular graph, say lattices in a two-dimensional plane, was applied to represent the relationship among factors in a real system. Till the end of 1950's, two Hungarian mathematicians, Paul Erdös and Alfréd Rényi, presented a random graph model or ER model. In the next forty years, it was believed by many researchers that the ER model was an optimum model to describe those real systems [15, 16]. In 1998, however, a significant breakthrough was made. Duncan J. Watts and Steven H. Strogatz [17] proposed a small-world network model (WS model) which is a new type of network between a regular lattice network and a random graph. This made researchers realize that most of real-world networks were neither purely regular networks nor purely random networks, but they were networks with statistical features that differed from two previously mentioned networks, so this new type of network was named Complex Network. WS model may well exhibit two features in a great number of social networks, i.e., small average path lengths and high clustering coefficient, which is also called Small-world Effect. Later, in 1999, Barabási and Albert [18] addressed a scale-free network model (BA model) by analyzing mass data of real-world networks, which was rooted in two generic mechanisms: Growth and Preferential attachment. So the notable property of this model was characterized by the evolution of the network, which was consistent with the feature of the real complex system.

There are three main statistical features to describe a complex network [19, 20]: (1) degree distribution, (2) average path length, (3) clustering coefficient.

(1) Degree distribution. The degree of a node in a network is the number of edges, which in some sense indicates whether it is important or not. The degree distribution is the probability distribution of these degrees over the whole network and $P(k)$ is the probability that a node chosen uniformly at random has degree $k$. There are two commonly observed degree distributions: (a) exponent distribution and (b) power-law distribution, $P(k) \sim k^{-\gamma}$.

(2) Average path length. The path length $d_{ij}$ between two nodes, $i$ and $j$, is defined as the number of edges on the shortest path, while the maximum of all path lengths in a network is called the diameter of the network $d = max_{i,j}(d_{ij})$. The average path length may be obtained by computing the mean of all shortest path lengths.

(3) Clustering coefficient. The clustering coefficient $C_i$ of a node $i$ describes the relationship among nodes that is connected to the node $i$. If there is a node $i$ with $k_i$ nearest neighbors in a network, there exist $k_i(k_i - 1)/2$ edges at best among its $k_i$ nearest neighbors. So the clustering coefficient may be defined as $C_i = 2E_i/k_i(k_i - 1)$, where $E_i$ represents really existing edges among its $k_i$ nearest neighbors. The clustering coefficient $C$ of the whole network is an average of the clustering coefficients $C_i$ of all nodes.

## 2.2 Algorithms using Flocking

Although flocking has been studied for years and applied to many fields, the idea of flocking dose not appear in the domain of data clustering until recent years. In 2002, G. Folino et al. [21] proposed an adaptive flocking algorithm, called SPARROW, for spatial data clustering. In their algorithm, a fixed number of agents were generated randomly in space, and then each agent chose his color according to the density of data in his neighborhood: red color, indicating an area with a high density; green color, a medium one; yellow color, a low one; white color, revealing an area without interesting patterns. The red and white agents stayed still in their positions as landmarks to signal the density of areas to the others, while the green and yellow agents moved in space in terms of Reynolds's three rules [22]: Cohesion, Separation, Alignment, where the green agents moved more slowly than the yellow agents. At the same time, new agents were regenerated to maintain the constant number of movable agents because both the red and white agents did not move and an agent would die when his period of life was end. Finally, if points in the circular neighborhood of red agents were not visited, a new cluster would be built; or if points in the area belonged to different clusters, they would be merged by red agents. In the next year, the idea of SNN algorithm [23] was added into SPARROW in order to build an algorithm called SPARROW-SNN [24]. In 2006, the P-SPARROW was constructed on P2P network [25]. Besides, Cui et al. [5] presented a flocking based algorithm for document clustering in 2006. In their algorithm, each document vector was mapped as a boid into a two-dimensional virtual space firstly. By means of four rules, Reynolds's three rules and an added *Feature Similarity and Dissimilarity rule*, and a set of preset weighted values, the velocity vector of a boid was computed in order to control the boid moving.

Unlike their algorithms, in our algorithm, each data point for clustering is regarded as a movable agent in space in order to improve the efficiency of clustering process, whereas movable agents are generated and added into the space of data points, and data points themselves do not move in Folino's SPARROW and its variants. Moreover, data points are clustered in a high-dimensional space instead of mapping them into a two-dimensional virtual space in Cui's algorithm. Further, a local potential function $\phi(\cdot)$ is employed by which a pair of data points interacts with one another. Thus, only two Reynolds's rules, Cohesion and Separation, are needed, when data points move. As a result, they gather automatically and establish some separating flocks. In addition, the relationship among data points is represented by a time-varying complex network that is formed by adding long-range connections for each data point after basic connections are set up among data points. The structure of the complex network provides a new topology for clustering, and it is more important that the weak ties built by long-range connections provide some hidden information for agents that cannot be perceived directly, which also bring about good results in experiments.

# 3 Proposed model

Assume a set $\boldsymbol{X}$ with $N$ movable agents, $\boldsymbol{X} = \{\boldsymbol{X}_1, \boldsymbol{X}_2, \cdots, \boldsymbol{X}_N\}$, which are distributed in a $m$-dimensional metric space and each of which represents a current position of an agent. In this metric space, a distance function $d : \boldsymbol{X} \times \boldsymbol{X} \longrightarrow \mathbb{R}$ is defined which satisfies that the closer the two agents are, the smaller the output is. Based on the distance function a distance matrix is computed whose entries are distances between any two agents. Thus, a weighted and directed $k$-nearest neighbors (knn) graph or network, $G(\boldsymbol{X}(t), E(t), d)$, is defined which represents the basic connections among all agents.

**Definition 1** *If there is a set $\boldsymbol{X}$ with $N$ movable agents, $\boldsymbol{X} = \{\boldsymbol{X}_1, \boldsymbol{X}_2, \cdots, \boldsymbol{X}_N\}$, the weighted and directed knn graph or network, $G(\boldsymbol{X}(t), E(t), d)$, is created as below.*

$$\begin{cases} \boldsymbol{X}(t) = \Big\{ \boldsymbol{X}_i(t), i = 1, 2, \cdots, N \Big\} \\ E(t) = \bigcup_{i=1}^{N} E_i(t) \\ E_i(t) = \Big\{ e\Big( \boldsymbol{X}_i(t), \boldsymbol{X}_j(t) \Big) \mid j \in \varphi_i(t) \Big\} \\ \varphi_i(t) = \Big\{ j \Big| j = \underset{h \in V}{argmink} \Big( \Big\{ d(\boldsymbol{X}_i(t), \boldsymbol{X}_h(t)), \boldsymbol{X}_h(t) \in \boldsymbol{X} \Big\} \Big) \Big\} \end{cases} \tag{1}$$

*where the function, $argmink(\cdot)$, is to find $k$ nearest neighbors of an agent which construct a neighbor set $\varphi_i(t)$.*

As agents move in space, their positions are constantly changing over time. So, each agent is exposed to a changing environment, which causes $k$ nearest neighbors of an agent different at almost all times. In other words, his edge set $E_i(t)$ and neighbor set $\varphi_i(t)$ are time-varying. As such, the shape of the weighted and directed knn graph $G(\boldsymbol{X}(t), E(t), d)$ is also changed over time.

Additionally, the average path length of the network will be reduced and the clustering coefficient will grow if long-range connections are added for each agent, so that a time-varying complex network is formed on the basis of the graph $G(\boldsymbol{X}(t), E(t), d)$.

**Definition 2** *Based on the weighted and directed knn graph $G(\boldsymbol{X}(t), E(t), d)$, a time-varying complex network $Z(\boldsymbol{X}(t), E'(t), d)$ is established by adding long-range connections.*

$$\begin{cases} \boldsymbol{X}(t) = \Big\{ \boldsymbol{X}_i(t), i = 1, 2, \cdots, N \Big\} \\ E'(t) = E(t) \bigcup \Big( \bigcup_{i=1}^{N} E_i''(t) \Big) \\ E_i''(t) = \Big\{ e\Big( \boldsymbol{X}_i(t), \boldsymbol{X}_j(t) \Big) \mid j \in \psi_i(t) \Big\} \\ \psi_i(t) = \Big\{ C^r \Big( \overline{\varphi_i(t)} \Big) \Big\} \\ \Gamma_i(t) = \varphi_i(t) \bigcup \psi_i(t) \end{cases} \tag{2}$$

*Here, $\psi_i(t)$ represents a long-range neighbor set of an agent $\boldsymbol{X}_i$; the long-range-connection-choosing (LRCC) function $C^r(\cdot)$ is to find $r$ other agents as long-range neighbors of the agent $\boldsymbol{X}_i$ outside his $k$ nearest neighbors and himself, and creates new edges connected to them; $\Gamma_i(t)$ represents a set of the agent $\boldsymbol{X}_i$'s all neighbors.*

Further assume that each agent establishes a local field at a position in space through a potential function $\phi(\cdot)$ that is a function of degree $Deg(\cdot)$ and distance $d(\cdot, \cdot)$. If an agent $\boldsymbol{X}_j$ is one of an agent $\boldsymbol{X}_i$'s neighbors, which means there is an edge connecting from the agent $\boldsymbol{X}_i$ to the agent $\boldsymbol{X}_j$, then the filed which is set up at the position of the agent $\boldsymbol{X}_i$ by the agent $\boldsymbol{X}_j$ may be computed according to the following formulation.

$$\phi\Big(\boldsymbol{X}_i(t), \boldsymbol{X}_j(t)\Big) \cdot \overrightarrow{n_{ij}} = \begin{cases} \frac{Deg_j(t)}{\left(d\left(\boldsymbol{X}_i(t),\boldsymbol{X}_j(t)\right)\cdot d\left(\boldsymbol{X}_i(0),\boldsymbol{X}_j(0)\right)\right)^2} \cdot \overrightarrow{n_{ij}} & \text{if } d\Big(\boldsymbol{X}_i(t), \boldsymbol{X}_j(t)\Big) > \theta \\ 0 & \text{otherwise} \end{cases} \tag{3}$$

where $\overrightarrow{n_{ij}} = \boldsymbol{X}_i(t) - \boldsymbol{X}_j(t)$ is a vector pointing radially to the agent $\boldsymbol{X}_j$; $d(\boldsymbol{X}_i(t), \boldsymbol{X}_j(t))$ and $d(\boldsymbol{X}_i(0), \boldsymbol{X}_j(0))$ represent the current distance and initial distance between the agent $\boldsymbol{X}_i$ and $\boldsymbol{X}_j$ respectively; the variable $\theta$ is a separating threshold which indicates that the agent $\boldsymbol{X}_i$ does not be acted by a field established by the agent $\boldsymbol{X}_j$ when the distance between them is no larger than the threshold $\theta$. In other words, for the agent $\boldsymbol{X}_i$, the agent $\boldsymbol{X}_j$ disappears in space temporally in this case.

All neighbors in the neighbor set $\Gamma_i(t)$ of the agent $\boldsymbol{X}_i$, $k$ nearest neighbors and $r$ long-range neighbors, set up fields at the position of the agent $\boldsymbol{X}_i$. As is known, fields follow a principle of linear superposition, so the total fields at the position of the agent $\boldsymbol{X}_i$ is the vector sum of individual fields.

$$\ddot{\boldsymbol{X}}_i(t) = \sum_{j \in \Gamma_i(t)} \phi\Big(\boldsymbol{X}_i(t), \boldsymbol{X}_j(t)\Big) \cdot \overrightarrow{n_{ij}} \tag{4}$$

Then, the agent $\boldsymbol{X}_i(t)$ will take a step of length $l_i$ along the direction of the total fields $\ddot{\boldsymbol{X}}_i(t)$ because all neighbors exert actions on him. However, we assume the agent $\boldsymbol{X}_i(t)$ moves so slowly that the velocities before and after he moves may remain zero, since all data points do not need to move in a direction as a whole group. Thus, the velocity becomes a constant, so one of Reynolds's three rules, Alignment, may be eliminated. Moreover, the moving distance of the agent $\boldsymbol{X}_i$ is proportional to the magnitude of the total fields $\ddot{\boldsymbol{X}}_i(t)$.

$$l_i = \alpha \cdot \Big|\ddot{\boldsymbol{X}}_i(t)\Big| = \alpha \cdot l'_i \tag{5}$$

Here, the variable $\alpha$ is a factor of proportionality which is computed by the below formulation.

$$\begin{aligned} \alpha &= \begin{cases} l_i^{ave}/l'_i & \text{if } l'_i/l_i^{ave} \\ 1 & \text{otherwise} \end{cases} \\ l_i^{ave} &= \frac{\sum_{j \in \varphi_i(t)} Deg_j(t) \cdot d\left(\boldsymbol{X}_i(t), \boldsymbol{X}_j(t)\right)}{\sum_{j \in \varphi_i(t)} Deg_j(t)} \end{aligned} \tag{6}$$

After each agent takes a step along the direction of the total fields, his position is updated according to Eq. (7).

$$\boldsymbol{X}_i(t+1) = \boldsymbol{X}_i(t) + \alpha \cdot \ddot{\boldsymbol{X}}_i(t) \tag{7}$$

When the positions of all agents in the system are updated, an iteration of the model is completed.

# 4 Algorithm and analysis

In this section, at first two different LRCC functions $(C_1^r(\cdot), C_2^r(\cdot))$ are designed, and then two clustering algorithms (FLCN1, FLCN2) based on them are established. Next, the clustering algorithms are elaborated and analyzed in detail.

## 4.1 Clustering algorithms

Assume an unlabeled dataset $\boldsymbol{X} = \{\boldsymbol{X}_1, \boldsymbol{X}_2, \cdots, \boldsymbol{X}_N\}$, whose each instance is with $m$ features. In the clustering algorithms, each data point in the dataset is considered as an agent who can move in space, and the relationship among all data points is represented by a time-varying complex network $Z(\boldsymbol{X}(t), E'(t), d)$.

Firstly, according to the proposed model, after a distance function $d : \boldsymbol{X} \times \boldsymbol{X} \longrightarrow \mathbb{R}$ is selected, a weighted and directed graph or network $G(\boldsymbol{X}(t), E(t), d)$ is constructed in terms of Def. 1. Then, when a long-range neighbor set $\psi_i(t)$ of an agent $\boldsymbol{X}_i$ is built by the LRCC function $C^r(\cdot)$, a time-varying complex network can be produced according to Def. 2. If different potential functions or LRCC functions are employed, neighbors in an agent's neighborhood and his moving direction and distance will be changed. In the end, there is no doubt that the obtained results are various. Here, we design two different LRCC functions $(C_1^r(\cdot), C_2^r(\cdot))$ and construct two clustering algorithms.

Algorithm FLCN1:

In Algorithm FLCN1, each data point selects long-range neighbors from all other data points except its $k$ nearest neighbors and himself, namely in a set $\{\overline{\varphi_i(t)}\}$, and then creates new connections, which is completed by a LRCC function $C_1^r(\cdot)$. Before long-range connections are created, the connecting probability of each data point is computed as below.

$$\omega_1^j = \frac{Deg_j(t) \cdot exp\Big(d\Big(\boldsymbol{X}_i(t), \boldsymbol{X}_j(t)\Big)\Big)^{-1}}{\sum_{j \in \{\overline{\varphi_i(t)}\}} Deg_j(t) \cdot exp\Big(d\Big(\boldsymbol{X}_i(t), \boldsymbol{X}_j(t)\Big)\Big)^{-1}} \tag{8}$$

The method of computing the probability is similar to that using in Ref. [26], but the degree of each data point is not involved in the computation of connecting probability in Ref. [26]. Next, those data points with the first to the $r$-th largest connecting probabilities will be found to form the long-range neighbor set of a data point $\boldsymbol{X}_i$.

$$\psi_i(t) = C_1^r\Big(\Big\{\overline{\varphi_i(t)}\Big\}\Big) = \underset{j \in \{\overline{\varphi_i(t)}\}}{argmaxr}\Big(\Big\{\omega_1^j, j \in \{\overline{\varphi_i(t)}\}\Big\}\Big) \tag{9}$$

where the function $argmaxr(\cdot)$ is to find the argument of the first to the $r$-th largest connecting probabilities.

Algorithm FLCN2:

In Algorithm FLCN1, when a data point chooses its $r$ long-range neighbors, $N-k-1$ other data points need to be checked. However, if the total number $N$ of data points is rather large, the LRCC function will search in a very large pool of data points, which causes much running time to be consumed. Nevertheless, the connecting probability of a data point is associated directly with its degree $Deg(\cdot) \in [k, N]$ and the reciprocal of distance between two data points $exp(d(\boldsymbol{X}_i(t), \boldsymbol{X}_j(t)))^{-1} \in [0, 1]$. Analyzing this carefully, we can find that the data points with large degrees are selected more easily, when the LRCC function is applied to find long-range neighbors for a data point. In conclusion, the clustering process is influenced strongly by data points with large degrees.

Based on the above fact, to reduce the search area, initially, $g = \eta \cdot N$ data points with large degrees are chosen to form the candidate long-range neighbor set $V$, where $\eta \in [0, 1]$ is a factor of proportionality and $g$ satisfies an inequality $r \leqslant g \ll N$.

$$V = \underset{j=\{1,2,\cdots,N\}}{argmaxg}\left(\left\{Deg_j(0), j \in \{1, 2, \cdots, N\}\right\}\right) \tag{10}$$

Here, the function $argmaxg(\cdot)$ is to find data points with the first to the $g$-th largest degrees to establish the candidate long-range neighbor set $V$. Next, only the connecting probability $\omega_2^j$ of each element in the set $V$ is computed.

$$\omega_2^j = \frac{Deg_j(0) \cdot exp\Big(d\Big(\boldsymbol{X}_i(t), \boldsymbol{X}_j(t)\Big)\Big)^{-1}}{\sum_{j \in \{\overline{\varphi_i(t)}\}} Deg_j(0) \cdot exp\Big(d\Big(\boldsymbol{X}_i(t), \boldsymbol{X}_j(t)\Big)\Big)^{-1}} \tag{11}$$

Finally, $r$ data points are selected from the candidate long-range neighbor set $V$ to form the long-range neighbor set $\psi_i(t)$ of a data point $\boldsymbol{X}_i \in \boldsymbol{X}$.

$$\psi_i(t) = C_2^r(V) = \underset{j \in V}{argmaxr}\left(\left\{\omega_2^j, j \in V\right\}\right) \tag{12}$$

Each data point is acted by the fields established by its $k$ nearest neighbors and $r$ long-range neighbors, and then takes a step along the direction of vector sum of all fields. When the sum of distances that all data points move is less than a preset threshold $\varepsilon$, this means the algorithm has converged, so the algorithm exits. At this time, we can observe that some separating parts emerge automatically in space, each of which corresponds to a cluster. The steps of two algorithms are summarized in Table 1.

## 4.2 Analysis of algorithm

In the course of data clustering, all neighbors in the neighbor set of a data point $\boldsymbol{X}_i$ including its $k$ nearest neighbors and long-range neighbors establish fields directed toward them at the position of $\boldsymbol{X}_i$. However, according to Eq. (3), if the distance between the data point $\boldsymbol{X}_i$ and one of its neighbors $\boldsymbol{X}_j$ is less than the separating threshold $\theta$, the neighbor $\boldsymbol{X}_j$ will not set up a field at the position of $\boldsymbol{X}_i$, i.e., the data point $\boldsymbol{X}_i$ will not be acted by the neighbor $\boldsymbol{X}_j$. As such, the data point $\boldsymbol{X}_i$ will move away from the neighbor $\boldsymbol{X}_j$ due to being acted together by all other neighbors, which can be viewed as Reynolds's Separation rule. On the other hand, each data point will move in the direction of the vector sum of fields that are produced by all neighbors, but the moving distance is proportional to the magnitude of the total field and cannot exceed the weighted mean of distances of its $k$ nearest neighbors. This is an efficient method to avoid that the moving distance of a data point is too large, because the actions that are exerted on the data point may be very strong in terms of Eq. (4) if the degrees of neighbors are large and the distances are small. Thus, the whole process makes data points approaching a center, which embodies Reynolds's Cohesion rule.

Table 1: Steps of clustering algorithm.

| |
|---|
| Select a distance function $d(\cdot,\cdot)$ |
| Set the number of long-range neighbors $r = k/2 \leq g$ and separating threshold $\theta$ |
| Repeat: |
| Produce the weighted and directed knn graph $G(\boldsymbol{X}(t), E(t), d)$ among data points according to Def. 1 |
| FLCN1: Build the long-range neighbor set $\psi_i(t)$ according to Eq. (9) |
| FLCN2: Build the long-range neighbor set $\psi_i(t)$ according to Eq. (11) |
| Add long-range connections to create a complex network $Z(\boldsymbol{X}(t), E'(t), d)$ according to Def. 2 |
| For each data point $\boldsymbol{X}_i \in \boldsymbol{X}$ |
| Compute the degree $Deg_i(t)$ of data point $\boldsymbol{X}_i$ |
| Compute the field in position $\boldsymbol{X}_i$ that is established by a neighbor in the neighbor set $\Gamma_i(t)$ according to Eq. (3) |
| Compute the vector sum $\ddot{\boldsymbol{X}}_i(t)$ of all fields in position $\boldsymbol{X}_i$ according to Eq. (4) |
| Compute the moving distance $l_i$ of data point according to Eq. (5) |
| End For |
| For each data point $\boldsymbol{X}_i \in \boldsymbol{X}$ |
| update the position of data point $\boldsymbol{X}_i$ according to Eq. (7) |
| End For |
| Until $\sum_{i=1}^{N} l_i < \varepsilon$ |

In the network, each data point is connected to several other data points with long distances by adding some long-range connections. Observing the network, we can find that the number of nodes that need to be visited from one to another is reduced, i.e., those added long-range connections shorten the average path length of the whole network. Besides, it is worth noting that data points with large degrees are selected and connected more easily, which can be derived from the formulation of the connecting probability, Eq. (8) or Eq. (11). Thus, the degrees of these data points will grow larger further, which makes the clustering coefficient of the network rising. Therefore, the network that is built by adding long-range connections for each data point on the basis of the weighted and directed knn graph is with the statistical features of the small-world complex network.

The long-range connections in the network may be also seen as certain weak ties [27] for those long-range neighbors only exert weak actions on a data point due to long distances between them according to Eq. (3). In some cases, however, these very weak ties greatly affect the evolution of the network. Just as mentioned in Ref. [27] and [28], when a person is hunting for a job, those who can provide useful information for him are not his close friends ($k$ nearest neighbors) but ones who are not familiar or met only one or two times (weak ties). This is because he and his close friends live in a same friend group and his friends are also friends. Thus, so much same information is shared in this friend group. Nevertheless, those who are not familiar with belong to other friend groups, so the information that they can give him is non-overlapped and more valuable for him.

After long-range connections are added into the network, weak ties similar to the above example are created among data points, which also provide some new information for each data point that he cannot notice when without long-range connections. This key point is exhibited clearly for data points in the boundary. Generally speaking, boundary points lie in areas of low density, which means there are long distances between them and their $k$ nearest neighbors. According to Eq. (4), the actions that it experiences are small, which causes its moving distance is also small for the distance is proportional to the total actions. By contrast, the actions that other data points experience are larger than that of boundary points, so their moving distances are larger too, which makes them approach some centers quickly. This process causes the situation of boundary points worse, because the distances between boundary points and their neighbors will be increased at the same time that both the actions and moving distances will be decreased further. However, if some long-range connections are added into the network, at beginning the actions that the data points experience will be larger, so the rate of converging to a center will be accelerated. It is more important that long-range connections provide some information about the distribution of densities of data points or positions of centroids for each data points since these long-range neighbors are often with large degrees.

# 5 Discussion

In the section, firstly, we discuss how the number of clusters is affected by the number $k$ of nearest neighbors changing. Then, for the Algorithm FLCN2, the relationship between the factor of proportionality $\eta$ and the clustering results are investigated, which provides a way to select the variable $\eta$. Finally, the rates of convergence of two clustering algorithms are compared.

## 5.1 Number of nearest neighbors vs. number of clusters

The number $k$ of nearest neighbors represents the number of neighbors to which a data point $\boldsymbol{X}_i \in \boldsymbol{X}$ connects. If the longest distance among the data point and its $k$ nearest neighbors is selected as a radius, then a virtual circle centered around the data point can be drawn. This circle may be viewed as the interaction range of the data point whose radius follows the increase of the number of nearest neighbors. For a dataset, the number $k$ of nearest neighbors determines the number of clusters in part. Generally speaking, the number of clusters decreases with the increase of the number of nearest neighbors. For example, if the number of nearest neighbors is small, the interaction range of a data point is small too. Further, considering connectivity of a graph, we can find that the interaction ranges of data points intersect each other slightly, so that the connected domain formed is also small. In the process of data points moving in space, they will be close to one another gradually, which causes both the interaction ranges of data points and the connected domain on the graph are decreased further. Hence, in this case, they gather together only with not-too-distance data points around them. As a consequence, all data points form many small clusters, as is shown in Fig. 1(a).

On the other hand, if the number of nearest neighbors is large, the interaction ranges of data points will be increased at the same time, which makes them intersect each other largely. Thus, the larger connected domains on the graph are formed. Even if the interaction ranges of data points are decreased due to data points converging to a center, the larger connected domains are established in contrast to that when selecting a small number of nearest neighbors. Finally, several big clusters are formed. Fig. 1 illustrates the relationship between the number of clusters and the number of nearest neighbors. As is analyzed above, eight clusters are obtained by the clustering algorithm, when $k = 8$. As the number of nearest neighbors rises, five clusters are obtained when $k = 15$, three clusters when $k = 20$. So, if the exact number of clusters is not known in advance, different number of clusters may be achieved by adjusting the number $k$ of nearest neighbors in practice.

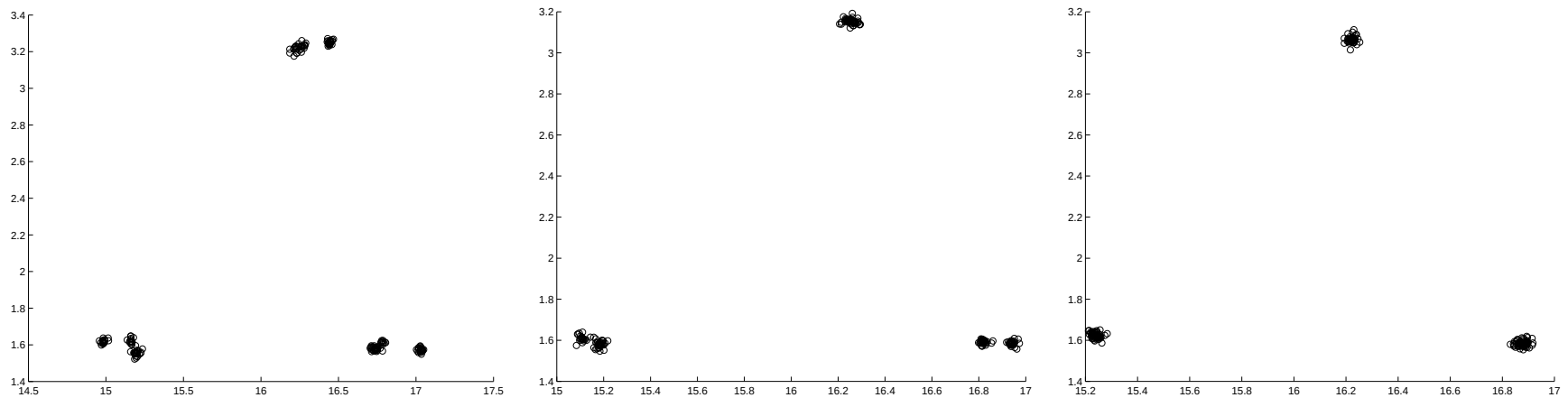

(a) $k = 8$ (b) $k = 15$ (c) $k = 20$

Figure 1: The number of nearest neighbors vs. number of clusters. (a) eight clusters are obtained, when the number of nearest neighbors is $k = 8$. (b) five clusters, when $k = 15$. (c) three clusters, when $k = 20$

## 5.2 Effect of the factor in Algorithm FLCN2

In Algorithm FLCN2, to simplify the computation, each data point does not choose its long-range neighbors from all other data points exclusive of its $k$ nearest neighbors, but from a candidate long-range neighbor set $V$ which is formed by a set of data points with large degrees selected beforehand. The number $g$ of elements in the set $V$, $g = \eta \cdot N$, depends on a factor of proportionality $\eta$ and satisfies $r \leq g \ll N$. For this reason, it is restricted in an interval $\eta \in [0, 0.5]$. For a same dataset, the results of Algorithm FLCN2 which are represented by the clustering accuracies (for definition see 6.2) are shown in Fig. 2 at the different number of nearest neighbors, when the factor of proportionality $\eta$ changes, i.e., the number of elements in the set $V$ changes.

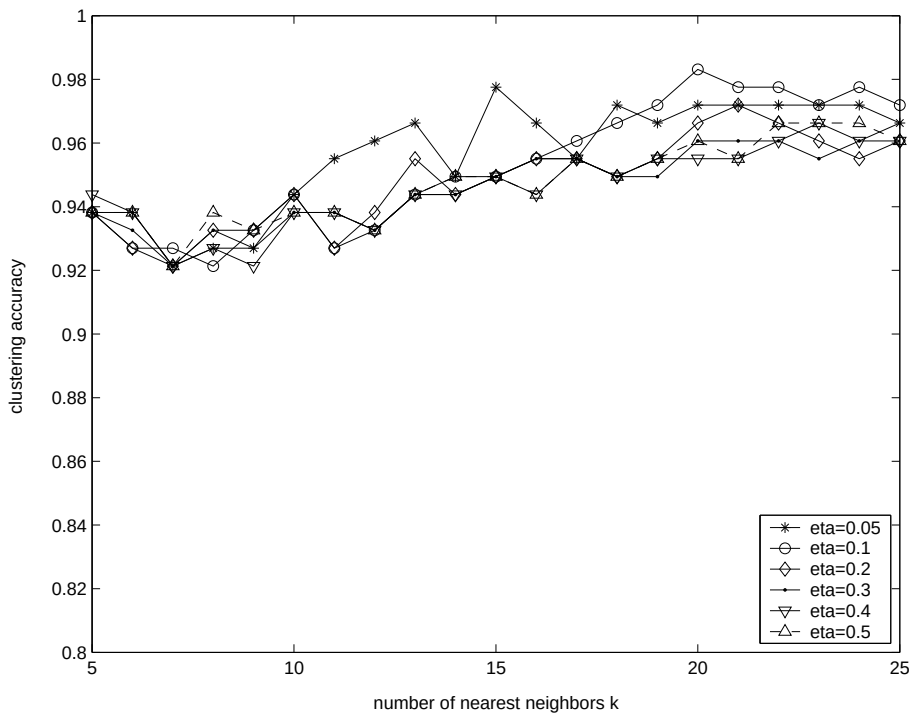

Figure 2: The factor of proportionality $\eta$ vs. results of Algorithm FLCN2.

From Fig. 2, we can see that the results fluctuate relatively largely when the variable is set at $\eta = 0.05$. This is because the elements in the candidate long-range neighbor set $V$ are only a few, and they may not be distributed uniformly in space but located at some places collectively. In this case, the effect of long-range connections cannot be demonstrated. However, when the variable $\eta$ takes other values, the obtained results seem stable relatively and at different $k$ similar results are achieved. While $\eta = 0.1$, the best result is achieved. As a whole, we recommend that the variable $\eta$ takes values in an interval $\eta \in [0.1, 0.2]$ for not only the inequality $g \ll N$ is satisfied, but also good results may be obtained due to enough elements in the set $V$. Therefore, in the later discussion, the variable $\eta$ in Algorithm FLCN2 takes $\eta = 0.1$.

## 5.3 Number of nearest neighbors vs. rates of convergence of algorithms

The rates of convergence of the proposed clustering algorithms are associated with the number of nearest neighbors closely too. As is known, the smaller the number of nearest neighbors is, the weaker the actions that a data point experiences are, and the smaller the magnitude of the total fields established by neighbors is according to Eq. (4). Again, the moving distance of the data point is proportional to the magnitude of the total fields, so it takes the data point more time to approach a center, which causes the rates of convergence of the algorithms slow down. On the other hand, if a big $k$ is adopted, the length of every step that a data point takes is increased, since it experiences stronger actions. As such, the rates of convergence of the algorithms grow. For a same dataset, the comparison of the rates of convergence about two algorithms is illustrated in Fig. 3, where every dot in Fig. 3 represents the sum of moving distances of all data points $\sum_{i=1}^{N} l_i$.

For each algorithm, the sum of first moving distances of all data points when $k = 5$ is obviously less than that when $k = 25$. As compared with Algorithm FLCN1, Algorithm FLCN2 only needs eight iterations to converge versus nine iterations of Algorithm FLCN1, when $k = 5$. While $k = 25$, two algorithms have converged after five iterations. In conclusion, as is shown in Fig. 3, the rates of convergence of two algorithms are fast enough.

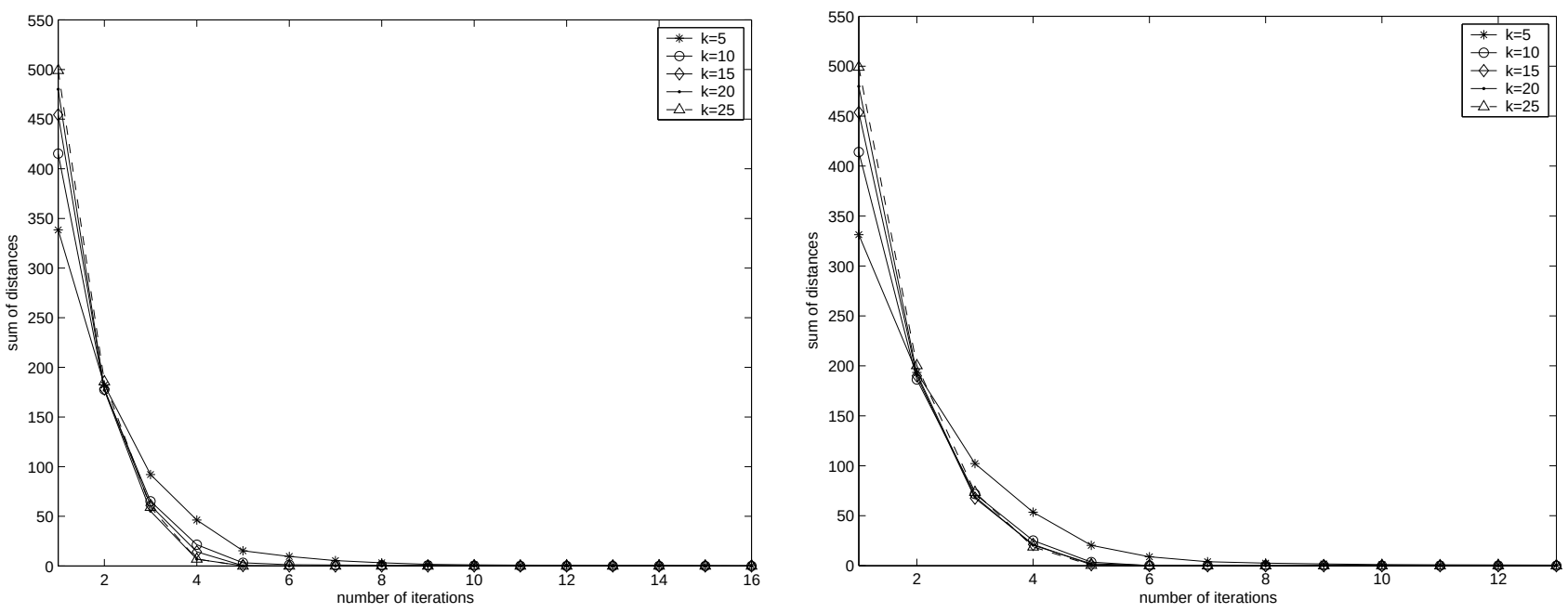

(a) Algorithm FLCN1 (b) Algorithm FLCN2

Figure 3: Comparison of rates of convergence of Algorithm FLCN1 and FLCN2.

# 6 Experiments

To evaluate these two clustering algorithms, we choose six datasets from UCI repository [29], which are soybean, Iris, Wine, Glass, Ionosphere and Breast cancer Wisconsin datasets, and complete the experiments on them. In this section, firstly we introduce these datasets briefly, and then demonstrate the experimental results.

## 6.1 Experimental setup

The original data points in above datasets all are scattered in high dimensional spaces spanned by their features, where the description of all datasets is summarized in Table 2. As for Breast dataset, those lost features are replaced by random numbers. Finally, this algorithm is coded in Matlab 6.5.

Table 2: Description of datasets.

| Dataset | Instances | Features | classes |
|---|---|---|---|
| Soybean | 47 | 21 | 4 |
| Iris | 150 | 4 | 3 |
| Wine | 178 | 13 | 3 |
| Glass | 214 | 9 | 6 |
| Ionosphere | 351 | 32 | 2 |
| Breast | 699 | 9 | 2 |

Throughout all experiments, data points in a dataset are considered as movable agents whose initial positions are taken from the datasets directly. Next, the weighted and directed knn graph representing basic connections among data points are created according to Def. 1, and then the complex network with long-range connections are constructed in terms of Def. 2, after a distance function is selected which only needs to satisfy that the more similar data points are, the smaller the output of the function is. In our experiments, the Euclidean distance function, 2-norm distance, is employed. In addition, the number of long-range neighbors is $r = k/2 \leq g$, and the separating threshold is $\theta = 0.1$.

## 6.2 Experimental results

Two clustering algorithms are experimented on the six datasets respectively. As is analyzed in section 5.1, for a dataset the number of clusters decreases with the increase of the number $k$ of nearest neighbors. Therefore, when a small $k$ is selected, it is possible that the number of clusters is larger than the preset number of clusters in the dataset, after the algorithm is end. So a merging-subroutine is called to merge unwanted clusters, which works in this way. At first, the cluster with the fewest data points is identified, and then is merged to the cluster whose distance between their centroids is smallest. This subroutine is repeated till the number of clusters is equal to the preset number. Moreover, the algorithms are run on every dataset at the different number of nearest neighbors. Those clustering results obtained by these two algorithms are compared in Fig. 4, in which each point represents a clustering accuracy.

**Definition 3** *$cluster_i$ is the label which is assigned to a data point $\boldsymbol{X}_i$ in a dataset by the algorithm, and $c_i$ is the actual label of the data point $\boldsymbol{X}_i$ in the dataset. So the clustering accuracy is [30]:*

$$\begin{aligned} accuracy &= \frac{\sum_{i=1}^{N} \lambda\Big(map(cluster_i), c_i\Big)}{N} \\ \lambda(map(cluster_i), c_i) &= \begin{cases} 1 & \textit{if } map(cluster_i) = c_i \\ 0 & \textit{otherwise} \end{cases} \end{aligned} \tag{13}$$

*where the mapping function $map(\cdot)$ maps the label got by the algorithm to the actual label.*

As is shown in Fig. 4, the similar results are obtained by two algorithms at different nearest neighbors, although the long-range neighbors are chosen from the candidate long-range neighbor set $V$ in Algorithm FLCN2. Additionally, we compare our results to those results obtained by other clustering algorithms, Kmeans [31], PCA-Kmeans [31], LDA-Km [31], on the same dataset. The comparison is summarized in Table 3.

Table 3: Comparison of clustering accuracies of algorithm.

| Algorithm | Soybean | Iris | Wine | Glass | Ionosphere | Breast |
|---|---|---|---|---|---|---|
| FLCN1 | 89.36% | 90% | 97.19% | 62.62% | 71.79% | 95.85% |
| FLCN2 | 91.49% | 90% | 98.31% | 63.08% | 71.79% | 95.85% |
| Kmeans | 68.1% | 89.3% | 70.2% | 47.2% | 71% | – |
| PCA-Kmeans | 72.3% | 88.7% | 70.2% | 45.3% | 71% | – |
| LDA-Km | 76.6% | 98% | 82.6% | 51% | 71.2% | – |

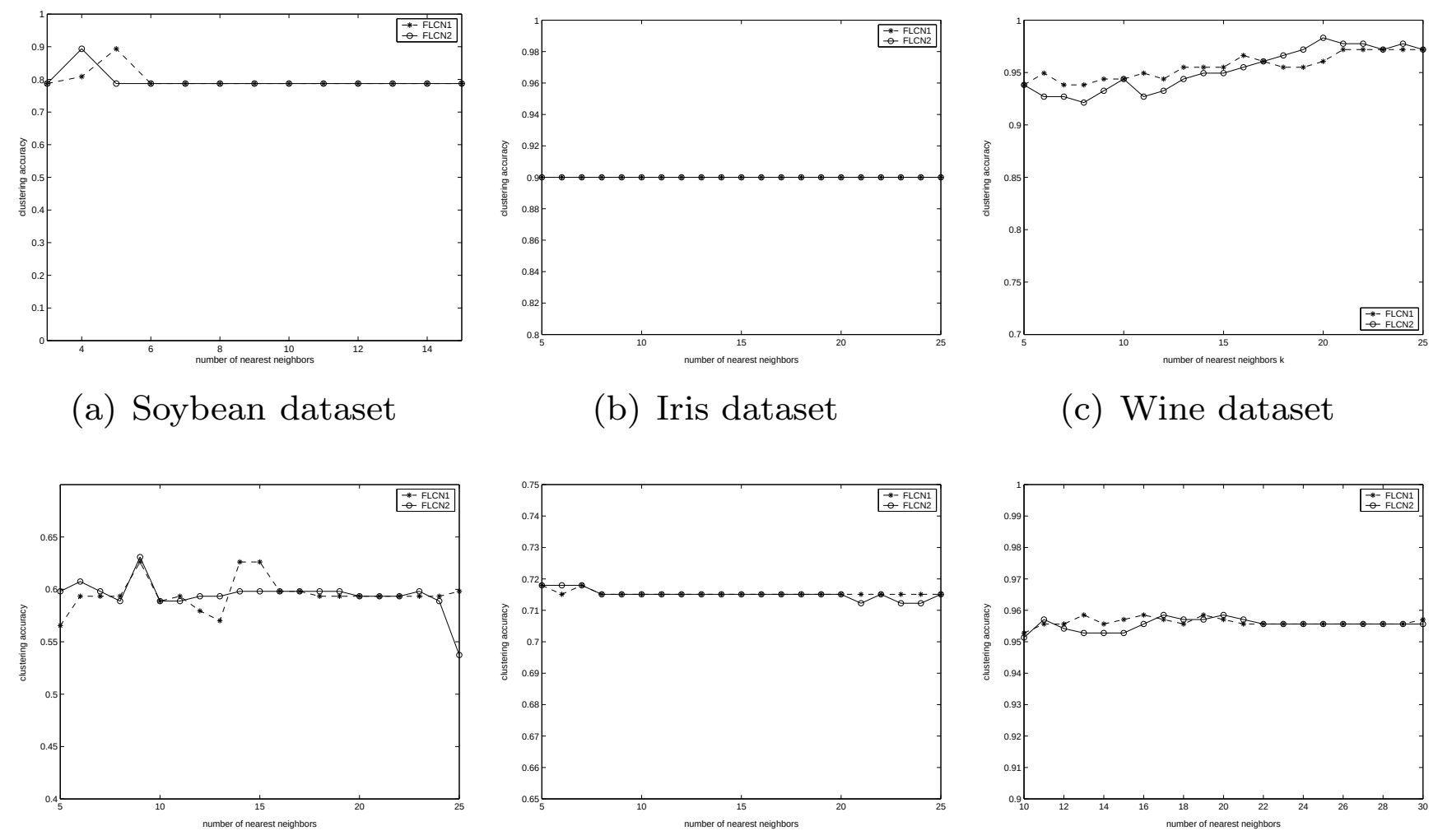

(a) Soybean dataset (b) Iris dataset (c) Wine dataset

(d) Glass dataset (e) Ionosphere dataset (f) Breast dataset

Figure 4: Comparison of clustering accuracies in Algorithm FLCN1 and FLCN2.

# 7 Conclusion

We have introduced a flocking model on the complex network, and developed two clustering algorithms based on it, in which data points in a dataset are considered as movable agents. When a distance function is selected, the basic connections among data points, a weighted and directed knn graph or network $G(\boldsymbol{X}(t), E(t), d)$, is created according to Def. 1. Next, the LRCC function $C^r(\cdot)$ is applied to add long-range connections for each data point. Then a time-varying complex network $Z(\boldsymbol{X}(t), E'(t), d)$ may be constructed in terms of Def. 2. By employing a potential function $\phi(\cdot)$ for each data point, at the position of a data point $\boldsymbol{X}_i$, the fields are established by all of its neighbors in the neighbor set $\Gamma_i(t)$. Thus, the data point $\boldsymbol{X}_i$ will take a step along the direction of vector sum of the total fields since it is acted by the fields. In the course of data points moving, the shape of the complex network will change over time. At beginning, the sum of moving distances of all data points is large, while the sum approaches a small value when the clusters are formed. If the sum is less than a preset threshold $\varepsilon$, the algorithm exits.

In Algorithm FLCN1, before a data point chooses its $r$ long-range connections, the connecting probabilities of all other data points except itself and its $k$ nearest neighbors are computed by a LRCC function $C_1^r(\cdot)$. Analyzing this process carefully, we can find that the selected long-range neighbors are usually with large degrees. Therefore, on the basis of this fact, in Algorithm FLCN2, initially $g$ data points with large degrees are selected to form a candidate long-range neighbor set $V$. As such, a data point only needs to choose its long-range neighbors from the set $V$ in order to reduce the search time. From the experimental results, two algorithms obtain similar results at a same $k$.

Those added long-range connections for each data point may be viewed as some weak ties in social networks because the distances between a data point and its long-range neighbors are so long that the actions that they exert on it are weak. However, generally speaking, long-range neighbors are usually with large degrees, so these weak ties provide some information of the distribution of data points with high densities for each data point. At the same time, they accelerate the rate that data points converge to some centers, as is demonstrated obviously for those boundary points. Besides, the proposed model provides a heuristic for clustering data points, as a consequence, data points belonging to the same class are close to each other, and form tight clusters, while the different clusters are away from one another.

In these two algorithms, when the exact number of clusters is unknown in advance, one can adjust the number of nearest neighbors to control the number of clusters which decreases with the increase of the number of nearest neighbors. We evaluate the clustering algorithms on six real datasets, experimental results have demonstrated that data points in a dataset are clustered reasonably and efficiently, and the rates of convergence of two algorithms are also fast enough. Additionally, these clustering algorithms can also detect clusters of arbitrary shape, size and density.

# Acknowledgments

The first author thanks Dr. Zhuo CHEN of Shanghai Jiao Tong University for helpful suggestions and discussions. This work is supported in part by the National Natural Science Foundation of China (No. 60405012, No. 60675055).